\documentclass{article}
\usepackage{spconf,amsmath,graphicx}
\usepackage{amssymb,amsfonts}
\usepackage{multirow}
\usepackage{url}
\usepackage{hyperref}
\usepackage{mathrsfs}
\usepackage{makecell}
\usepackage{xcolor}
\usepackage{arydshln}

\makeatletter
\def\endthebibliography{%
	\def\@noitemerr{\@latex@warning{Empty `thebibliography' environment}}%
	\endlist
}
\makeatother

\title{Cross-target Stance Detection by Exploiting Target Analytical Perspectives}
\name{Daijun Ding$^{1}$, Rong Chen$^{1}$, Liwen Jing$^2$, Bowen Zhang$^{3}$\sthanks{Corresponding author (Bowen Zhang: zhang\_bo\_wen@foxmail.com). \\ This work was supported in part by the National Nature Science Foundation of China (No.62306184 and No.62002122), and the Natural Science Foundation of Top Talent of SZTU (grant No. GDRC202320), and the Research Promotion Project of Key Construction Discipline in Guangdong Province (No. 2022ZDJS112).},  Xu Huang$^4$, Li Dong$^{3}$, Xiaowen Zhao$^5$, Ge Song$^6$ }
\address{
$^1$College of Applied Science, Shenzhen University, Shenzhen, China\\
 $^2$ Shenzhen X-Institute, Shenzhen, China\\
	$^3$College of Big Data and Internet,
	 Shenzhen Technology University,
	 Shenzhen, China\\
	 $^4$Computer Science \& Technology, Harbin Institute of Technology, Shenzhen, China\\
      $^5$Aoyama Gakuin University, Tokyo, Japan\\
	 $^6$College of Mathematics and Informatics, South China Agricultural University, Guangzhou, China
 }
\begin{document}
%
\maketitle
\begin{abstract}
Cross-target stance detection (CTSD) is an important task, which infers the attitude of the destination target by utilizing annotated data derived from the source target. 
One important approach in CTSD is to extract domain-invariant features to bridge the knowledge gap between multiple targets.
However, the analysis of informal and short text structure, and implicit expressions, complicate the extraction of domain-invariant knowledge.
In this paper, we propose a Multi-Perspective Prompt-Tuning (MPPT) model for CTSD that uses the {analysis perspective} as a bridge to transfer knowledge. 
First, we develop a two-stage instruct-based chain-of-thought method (TsCoT) to elicit target analysis perspectives and provide natural language explanations (NLEs) from multiple viewpoints by formulating instructions based on large language model (LLM).
Second, we propose a multi-perspective prompt-tuning framework (MultiPLN) to fuse the NLEs into the stance predictor.
Extensive experiments results demonstrate the superiority of MPPT against the state-of-the-art baseline methods.
\end{abstract}
\section{Introduction}\label{Introduction}

The goal of stance detection is to automatically predict the attitude (i.e., \textit{favor}, \textit{against}, or \textit{neutral}) of an opinionated text towards a given target \cite{augenstein2016stance}.
Remarkable results have been attained in stance detection when abundant labeled samples are accessible \cite{du2017stance}. 
However, creating data with rich annotations is a labor-intensive and time-consuming endeavor.
Consequently, the exploration of cross-target stance detection (CTSD), which deduces the stance towards the destination target by leveraging annotated data derived from the source target, is gaining widespread attention  \cite{zhu2022enhancing, DanZCBH22}.
Typically, the targets are within the same domain \cite{bowenacl}.


So far, extracting domain-invariant features to address the knowledge gap between multiple targets is an important method in CTSD \cite{xu2018cross, wei2019modeling,liang2022zero, zhang2022would}.
However, applying these approaches to stance detection in practical scenarios poses significant challenges. 
These challenges include the analysis of informal and brief social media texts, as well as the prevalence of implicit expressions of stance towards the given target, both of which complicate the extraction of domain-invariant knowledge \cite{bowenacl}.

To alleviate the aforementioned issues, we propose a Multi-Perspective Prompt-Tuning (MPPT) model for CTSD, which leverages the shared analysis perspectives to transfer knowledge between targets.
The proposed model is predicated on the notion that humans tend to approach the analysis of different targets within the same domain from similar perspectives, which may constitute domain-invariant knowledge and thus facilitate knowledge transfer between source and target domains \cite{zhang2023investigating}. 
For instance, when assessing the stances of both ``Hillary Clinton'' and ``Donald Trump,'' one might consider perspectives such as political ideology and personal characteristics. 
These shared analytical angles can serve as a bridge, allowing for knowledge transfer across different targets.
Specifically, we first propose a two-stage instruct-based chain-of-thought method (TsCoT).
TsCoT automatically elicits target analysis perspectives and provides natural language explanations (NLEs) from multiple viewpoints by formulating instructions based on LLM.
In addition, we design an attention-based prompt-tuning network (MultiPLN) that effectively integrates NLEs to enhance the accuracy of stance detection.

The main contributions of this paper can be summarized as follows.
(1) To the best of our knowledge, this is the first work that leverages the shared analysis perspectives to transfer knowledge between targets for stance detection. This is the most significant contribution of our study.
(2) We develop MPPT, which contains TsCoT and MultiPLN, for eliciting and integrating target analysis perspective and NLEs, respectively.
(3) Extensive experiments are conducted on widely used datasets to verify the effectiveness of MPPT for CTSD. The experimental results show that MPPT consistently outperforms other methods. 

\begin{figure}[htbp]
\centering
\includegraphics[width=0.95\linewidth]{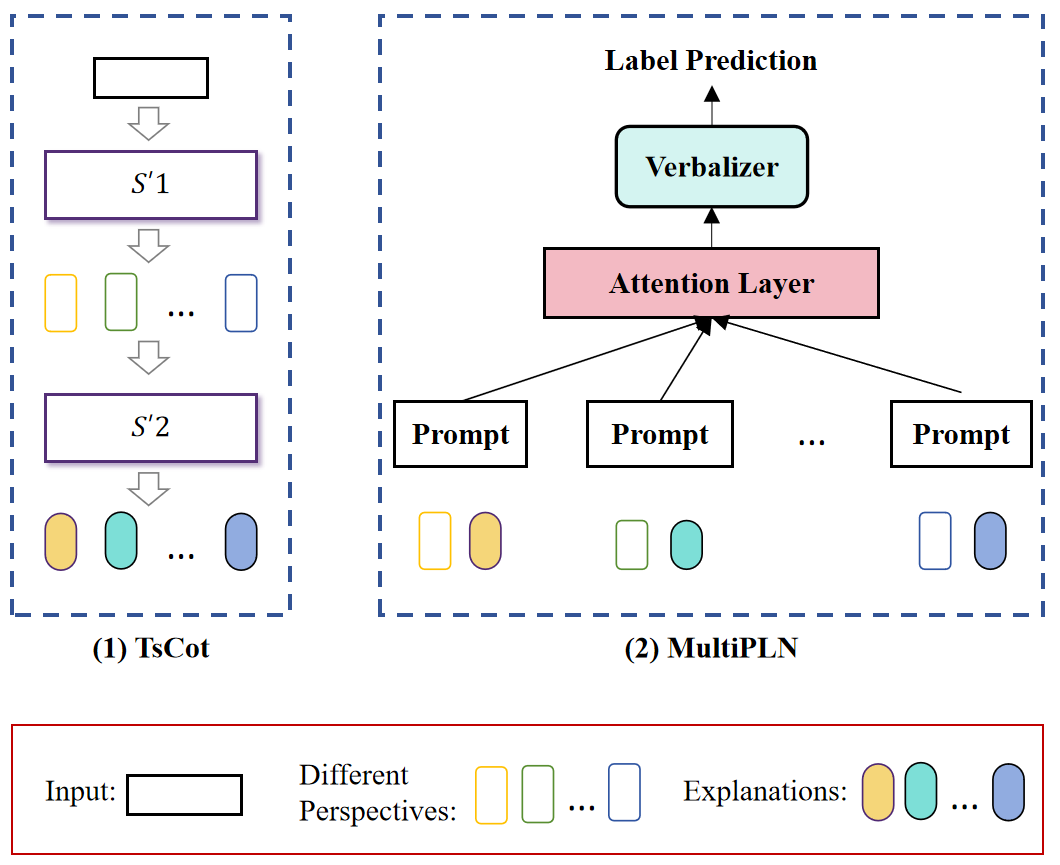}
\caption{The overall architecture of MPPT. \label{fra}}
\end{figure}




\section{Our Methodology}

We use $X^{s} = \{x^{s}_{i}, q^{s}_{i}\}$ to denote the collection of labeled data in the source domain, where each $x$ denotes the input text and $q$ denotes the corresponding target. Each sentence-target pair ($x^{s},q^{s}$) has a stance label $y^{s}$.  
Given an input $x^t$ and a target $q^t$ in the target domain, this study aims to predict a stance label by using the model learned with the labeled data in source domain.  
As illustrated in Fig. 1, MPPT consists of TsCoT and MultiPLN. 
Next, we will introduce the main components of MPPT in detail.

\subsection{TsCoT} 
TsCoT is proposed to elicit target analysis perspectives and provides natural language explanations (NLE) by formulating instructions based on LLM.
Specifically, TsCoT consists of the following steps:

\textbf{Step 1}: We first feed the constructed instruction $S'1$ into the LLM to elicit the view of the prediction of LLM. Here, $\gamma$ is a hyperparameter that varies within the range of 2 to 8.

\begin{center}
\fcolorbox{black}{blue!10}{\parbox{0.95\linewidth}{$S'1$:
From what angles do you think people might state their stance on the [given target: $q$]. List the $\gamma$ angles you can think of.}}
\end{center}


\textbf{Step 2}: 
After obtaining $\gamma$ perspectives, we let the LLM generate the stance analysis explanations for tweets under each perspective, which is represented by $k$.

\begin{center}
\fcolorbox{black}{blue!10}{\parbox{0.95\linewidth}{$S'2$:
Oriented to the [given target: $q$], given the input [given input: $x$], and under the [given perspective: $i \in \gamma$], give the stance analysis thinking or explanation. Give a positional judgment (favor,against,none) at the end.
}}
\end{center}

\subsection{MultiPLN}
\textbf{Preliminary: Prompt-tuning.}
Prompt-tuning is a transformative approach that reframes stance detection as a masked language modeling task. 
Specifically, prompt-tuning adopts a text template $p$ which is incorporated into the given text $x$ and the target $q$.
Formally, we denote the packed input as:
``$x_{p}$ = x. The attitude to $q$ is {[MASK]}''.
Let M be the LLM, which provides the probability of each word $v$ in the vocabulary being filled in [MASK] given $P_M($[MASK]$ = v|x_{p})$.
Here, $v$ represents the defined label word in the verbalizer. To map the probabilities of these words to the probabilities of the labels, a verbalizer is utilized as a mapping function $f$ from the defined words in the vocabulary, which form the label word set $V$, to the label space $Y$, i.e., $f: V \to Y$. 
Formally, the probability $P(y|x_p)$ of
label $y$, is computed as:
\begin{equation}
\label{2}
P(y|x_p)= \mu(P_M(\text{[MASK]} = v | x_p) | v \in V).
\end{equation}
where  $\mu$ serves as a crucial component in transforming the probability distribution over label words to the probability distribution over labels. 

\textbf{Prompt design.}
The key to a prompt-based method for stance detection is to construct the appropriate prompt. 
To effectively utilize background explanation knowledge $k$, our approach involves the creation of multiple prompts ($x_p$) derived from varied perspectives.

\begin{center}
\fcolorbox{black}{yellow!10}{\parbox{0.95\linewidth}{Template ($x_p$): [Given input: $x$]. From the perspective of [given perspective: $i \in \gamma$] and [given explanations: $k_i$]. The attitude to [given target: $q$] is $[$MASK$]$.
}}
\end{center}

\textbf{Attention layer.}
After constructing the template, we utilize a novel solution that involves the mapping of labels onto continuous vectors, called stance vectors, instead of explicit words or phrases. 
Specifically, we feed each $x_p^i$ into a LLM $M$ to acquire $r_i$.
Then, our method revolves around the generation of $\gamma$ distinct vectors ($R = \{r_i\}^\gamma$) that correspond to the ones generated by the [MASK] of each template $x_p$.
These vectors can be trained during optimization.
To ensure coherence with token embeddings within the pre-trained language model (PLM), the stance vectors have been dimensionally aligned with the size of hidden embeddings.
Subsequently, we use a randomly initialized learnable representation $h$ as the attention query to compute the attention weight $\alpha_t$ for the $t$-th word:
\begin{gather}
\alpha _t = softmax(h^\mathrm{T} R)\\
\mathit{e} = \sum_{t=1}^{n} \alpha_t R\ .
\end{gather}

\subsection{Stance Classification}
We classify the stance expressed in the text by assessing the semantic similarity between the target-aware stance vectors and the average of the label vector (which is defined in the verbalizer).
The process of predicting masked words in a prompt-based context is a multifaceted task, as there may be several suitable words that correspond to the linguistic context.
Most existing methods mainly focus on leveraging limited information to construct the verbalizer.
To improve the accuracy of stance detection, we introduce SenticNet \cite{cambria2022senticnet} as prior knowledge to expand the label words.
SenticNet can obtain its related semantically related words according to the given word.
For example, the semantic-related words of ``favor'' from SenticNet are ``happily, pleased, agree, affirmative''.
Specifically, to utilize semantic knowledge, we send the word in candidate (i.e., favor, against, none.) into the SenticNet and acquire the semantic-related words to define the verbalizer.
Here, the embedding of semantic-related word is defined as $w$.

Formally, based on the words provided by the verbalizer, we calculate the probability of selecting token $w$ as the label word.

\begin{gather}
\delta = \frac{\exp{(w_i \cdot \mathit{e})}}{\sum_{w_j \in V} \exp{(w_j \cdot \mathit{e}})} \ ,
\end{gather}
where $w$ is the embedding of the token in the verbalizer.
Then we sum the words' probabilities of each label, which denotes as $\hat{y}$.
Finally, the loss function can be effectively implemented through the utilization of the standard cross-entropy method.




\section{Experiments}
\subsection{Experimental Setup}
\textbf{Experimental Data}.
{SEM16} \cite{StanceSemEval2016} contains four targets: \textit{Donald Trump} (D), \textit{Hillary Clinton} (H), \textit{Legalization of Abortion} (L), and \textit{Feminist Movement} (F). 
Following \cite{bowenacl,liang2022jointcl, zhu2022enhancing}, we construct four CTSD tasks (\textit{F$\to$L, L$\to$F, H$\to$D, D$\to$H}). Here, the left side of the arrow corresponds to the source target and the right side of the arrow denotes the destination target.
{VAST} \cite{allaway2020zero} is a zero-shot stance detection (ZSSD) dataset with 4003 training topics and separate dev and test sets.

\textbf{Compared Baseline Methods}.
We assess and contrast our model against a number of reliable baselines, as follows:
{BiCond} \cite{augenstein2016stance}, CrossNet \cite{xu2018cross}, SEKT \cite{bowenacl}, {BERT-FT} \cite{BERT}, {PT-HCL} \cite{liang2022zero}, JointCL \cite{liang2022jointcl}, 
TGA-Net\cite{allaway2020zero}, CKE-Net\cite{liu2021enhancing}, {MPT \& KEPROMPT} \cite{huang2023knowledge}, 
{TPDG} \cite{LiangF00DHX21}, TarBK \cite{zhu2022enhancing}.



\textbf{Evaluation Metrics and Implementation Details}.
For SEM16 dataset, following \cite{liang2022zero} we utilize the Macro-avg: average of F1 on $Favor$ and $Against$. For VAST dataset, following \cite{li2023tts}, we employ Macro F1: average of each label. To evaluate the stability of the model. 
In the experiments, we select the BERT-base uncased as the skeleton of prompt-tuning framework, and  GPT-3.5-0301 for TsCoT.
The AdamW optimizer is applied to train the model, with a mini-batch size of 32 and a learning rate of 2e$^{-5}$.

\begin{table}[htbp]
\begin{center}	
\small
	\resizebox{0.85\linewidth}{!}{
\begin{tabular}{c|l|cccc}
\hline 
&Methods & F$\to$L  & L$\to$F & H$\to$D & D$\to$H \\ 
\hline 
\multirow{4}{*}{Sta.}&BiCond        & 45.0   & 41.6 & 29.7 & 35.8  \\
&CrossNet       & 45.4  & 43.3 & 43.1   & 36.2  \\
&SEKT            & 53.6 & 51.3 &  47.7   &42.0  \\
&TPDG            & 58.3  & 54.1 & 50.4 & 52.9   \\ 
\cdashline{1-6}[2pt/3pt]
\multirow{9}{*}{BERT}&BERT-FT            & 47.9  & 33.9 & 43.6 & 36.5\\    
&MPT  & 42.1  & 47.6  & 47.1  &  58.7    \\
&KEPROMPT    & {49.1} & 54.2 &  54.6   & 60.9  \\
&JointCL & 58.8  & 54.5& 52.8 &54.3 \\
&PT-HCL & 59.3 & 54.6 & 53.7 & 55.3\\
&TarBK &  59.1&  54.6&   53.1&  54.2 \\
\cdashline{2-6}[2pt/3pt]
&MPPT           & \textbf{67.4}$^\dag$ & \textbf{66.9}$^\dag$ &  \textbf{67.2}$^\dag$   & \textbf{73.5}$^\dag$  \\  
&\ \ - w/o T  &  52.7  & 42.2  &  47.7   &  53.1  \\  
&\ \ - w/o S  &  63.5  & 62.8  &  65.6  &  70.8  \\  
\hline		
	\end{tabular}
	}
\caption{Performance comparison of CTSD of Macro-F1. The $^\dag$  mark refers to a $p$-value $<$ 0.05. 
The best scores are in bold. 
}
\label{allll}
\end{center}
\end{table}

\begin{table}[htbp]
	\begin{center}
	\small
		\resizebox{0.65\linewidth}{!}{
	\begin{tabular}{c|l|ccc}
	\hline
&\multirow{2}{*}{Methods} & \multicolumn{3}{c}{VAST}
		\\ \cline{3-5}
		  && Con & Pro & All  \\
	\hline

\multirow{4}{*}{Sta.} & BiCond          & 47.4 & 44.6  & 42.8    \\
&CrossNet          & 43.4 & 46.2  & 43.4    \\
&SEKT          & 44.2 & 50.4  & 41.8    \\
&TPDG          & 49.6 & 53.7  & 51.9    \\
	\cdashline{1-5}[2pt/3pt]

\multirow{5}{*}{BERT}&BERT-FT   & 58.4 & 54.6 & 66.1   \\ 
&TGA-Net  &   61.7  &  63.4  & 64.7 \\
&CKE-Net  & 62.1 & 62.9 & 69.2  \\
&JointCL    & 63.3 & 63.5 & 71.3 \\ 

   	\cdashline{2-5}[2pt/3pt]
&MPPT  & \textbf{69.2}$^\dag$ & \textbf{67.6}$^\dag$ & \textbf{74.4}$^\dag$  \\
	\hline 	
	\end{tabular}
	}
\caption{Performance comparison of ZSSD.}
	\label{intarget}
\end{center}
\end{table}

\begin{table*}[htbp]
\begin{center}	
	\resizebox{\linewidth}{!}{
\begin{tabular}{l|c|c}
\hline 
Target & Donald Trump & Hillary Clinton \\
\hline
Tweet & \makecell{
Donald Trump gets the boot from \#NBC for calling immigrants rapists and murderers. Hey\\  
Donnie, where's your wife from again?\textit{[Against]}} 
& 
\makecell{
Hilary has lied, deleted Benghazi emails, and betrayed the trust of Americans scandal\\
after scandal. \textit{[Against]}} \\  \hline
Perspective & Personal characteristics & Personal characteristics \\ 
\cdashline{1-3}[2pt/3pt] 
NLEs & \makecell[l]{
... this input portrays Donald Trump as someone who is insensitive and disrespectful towards\\ immigrants, and possibly hypocritical in his views on immigration. ...
}
&
\makecell[l]{
... The input portrays Hillary Clinton as someone who has lied, deleted Benghazi emails, and\\ betrayed the trust of Americans scandal after scandal. This negative portrayal of her personal\\ characteristics suggests that she may not be trustworthy and may have a tendency to deceive people. ... } \\
\hline
Perspective & Political ideology & Political views \\
\cdashline{1-3}[2pt/3pt] 
NLEs & \makecell[l]{
... From a political ideology angle, this input suggests a critique of Trump's conservative\\
and nationalist views on immigration. ... } 
&
\makecell[l]{
... From a political views angle, this could be seen as a negative reflection on her character\\
and ability to lead. ...
 } \\ 
\hline
	\end{tabular}
	}
\caption{Case study of perspectives and NLEs.}
\label{allll}
\end{center}
\end{table*}

\subsection{Overall Performance} 

We report the main experimental results of ZSSD on four CTSD tasks in Table 1. 
We observe that our MPPT performs consistently better than all baseline models on all four tasks, which verifies the effectiveness of our proposed approach in CTSD. 
Further, the significance tests of our MPPT over comparison methods show that our MPPT presents a statistically significant improvement in terms of most evaluation metrics  (with $p$-value $<$ 0.05).
For example, our method improves 13.0\% over the best competitor (PT-HCL) on average for CTSD tasks.
The state-of-the-art performance comes from its two characteristics: (i) we develop a TsCoT to fully exploit the explanation for different analysis perspectives to the target; (ii) a MultiPLN is proposed to better fuse the external explanation knowledge for different perspectives.
In addition, our model achieves better performance than the knowlege-enhanced model (TarBK), which demonstrates that introducing the analysis perspective to bridge the knowledge gap is more effective than structural background knowledge. 
Compared with the prompt-tuning based model (KEPROMPT), our model achieves significant improvement on all datasets, which indicates that leveraging background knowledge is more effective to help LLM learn stance information.
We further conduct comparison experiments in ZSSD, a special form of CTSD. The challenge of this task is heightened due to the destination target being unseen during training. 
Experimental results are documented in Table 2.
Note our MPPT framework achieves consistently better performance on all zero-shot conditions, which verifies that our MPPT can generalize the superior learning ability to zero-shot scenario.

\begin{figure}[htbp]
	\centering
	\includegraphics[width=1\linewidth]{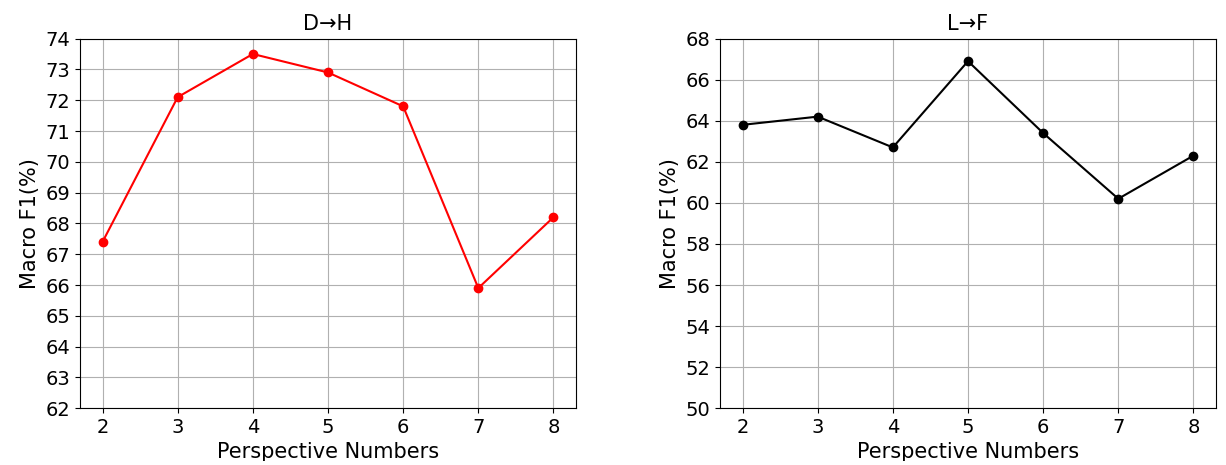}
\caption{Performance of perspective numbers.}
 \label{agu1}
\end{figure}

\subsection{Ablation Study}
To investigate the impact of each part on our model, we perform the ablation test by discarding TsCoT (denoted as ``- w/o T'') and SenticNet (denoted as ``-w/o S''). 
We report the results in Table 1.
We observe that the removal of TsCoT sharply degrades the performance, which verifies the effectiveness and significance of utilizing the analysis perspective as the bridge across domains.
Furthermore, integrating SenticNet to augment the verbalizer effectively boosts the prediction accuracy of the prompt learning framework.
Not surprisingly, combining all factors achieves the best performance for all the experiments.





\textbf{Analysis of Perspective.}
The number of perspectives is one of the most important hyperparameters, which has a big influence on the performance of MPPT. We run the experiments with the number of perspectives per target from 2 to 8, the results shows in Fig. 2. Taking D$\to$H as examples, we investigate that the accuracy improves as the perspective number increases. After 4 perspectives, the accuracy will be decreased.
This phenomenon is attributable to TsCoT's elicitation of four highly correlated perspectives from both D and H. 
As a result, the model culminates in its best performance when operating with these four viewpoints. Nevertheless, an increase in the number of perspectives leads to the introduction of conflicting information, thereby inducing noise and effectuating a deterioration in the model's accuracy.

\subsection{Case Study} 
We perform a case study, which MPPT accurately predicted, as illustrated in Table 3.
First, we observed that the two perspectives, Personal characteristics and Political ideology (views), are basically the same. Therefore, these perspectives serve as target-invariant information and can help the model learn transferable knowledge.
Second, NLEs under the same perspective can help the model learn knowledge shared by different targets to help improve the accuracy of cross-target prediction.  
For example, from the perspective of personal characteristics, both targets concurrently describe being unreliable and untrustworthy, which contributes to the correct prediction.
We maintain that this is the pivotal factor in augmenting CTSD performance.




\section{Conclusion}

In this paper, we proposed a MPPT for CTSD model that uses the {analysis perspective} as a bridge to transfer knowledge. 
Specifically, we first proposed a TsCoT to elicit target analysis perspectives and provide NLE from multiple viewpoints based on LLM.
Second, we developed a MultiPLN to fuse the NLE into the stance predictor.
Extensive experiments results demonstrate the superiority of our model against the state-of-the-art baseline methods.



\bibliographystyle{IEEEbib}
\bibliography{refs}

\begin{thebibliography}{10}

\bibitem{augenstein2016stance}
I~Augenstein, T~Rocktaeschel, A~Vlachos, and K~Bontcheva,
\newblock ``Stance detection with bidirectional conditional encoding,''
\newblock in {\em Proceedings of the Conference on Empirical Methods in Natural
  Language Processing}. Sheffield, 2016.

\bibitem{du2017stance}
Jiachen Du, Ruifeng Xu, Yulan He, and Lin Gui,
\newblock ``Stance classification with target-specific neural attention
  networks,''
\newblock International Joint Conferences on Artificial Intelligence, 2017.

\bibitem{zhu2022enhancing}
Qinglin Zhu, Bin Liang, Jingyi Sun, Jiachen Du, Lanjun Zhou, and Ruifeng Xu,
\newblock ``Enhancing zero-shot stance detection via targeted background
  knowledge,''
\newblock in {\em Proceedings of the 45th International ACM SIGIR Conference on
  Research and Development in Information Retrieval}, 2022, pp. 2070--2075.

\bibitem{DanZCBH22}
Yuhao Dan, Jie Zhou, Qin Chen, Qingchun Bai, and Liang He,
\newblock ``Enhancing class understanding via prompt-tuning for zero-shot text
  classification,''
\newblock in {\em {IEEE} International Conference on Acoustics, Speech and
  Signal Processing, {ICASSP} 2022, Virtual and Singapore, 23-27 May 2022}.
  2022, pp. 4303--4307, {IEEE}.

\bibitem{bowenacl}
Bowen Zhang, Min Yang, Xutao Li, Yunming Ye, Xiaofei Xu, and Kuai Dai,
\newblock ``Enhancing cross-target stance detection with transferable
  semantic-emotion knowledge,''
\newblock Association for Computational Linguistics, 2020.

\bibitem{xu2018cross}
Chang Xu, Cecile Paris, Surya Nepal, and Ross Sparks,
\newblock ``Cross-target stance classification with self-attention networks,''
\newblock in {\em Proceedings of the 56th Annual Meeting of the Association for
  Computational Linguistics (Volume 2: Short Papers)}, 2018, pp. 778--783.

\bibitem{wei2019modeling}
Penghui Wei and Wenji Mao,
\newblock ``Modeling transferable topics for cross-target stance detection,''
\newblock in {\em Proceedings of the 42nd International ACM SIGIR Conference on
  Research and Development in Information Retrieval}. ACM, 2019, pp.
  1173--1176.

\bibitem{liang2022zero}
Bin Liang, Zixiao Chen, Lin Gui, Yulan He, Min Yang, and Ruifeng Xu,
\newblock ``Zero-shot stance detection via contrastive learning,''
\newblock in {\em Proceedings of the ACM Web Conference 2022}, 2022, pp.
  2738--2747.

\bibitem{zhang2022would}
Bowen Zhang, Daijun Ding, and Liwen Jing,
\newblock ``How would stance detection techniques evolve after the launch of
  chatgpt?,''
\newblock {\em arXiv preprint arXiv:2212.14548}, 2022.

\bibitem{zhang2023investigating}
Bowen Zhang, Xianghua Fu, Daijun Ding, Hu~Huang, Yangyang Li, and Liwen Jing,
\newblock ``Investigating chain-of-thought with chatgpt for stance detection on
  social media,''
\newblock {\em arXiv preprint arXiv:2304.03087}, 2023.

\bibitem{cambria2022senticnet}
Erik Cambria, Qian Liu, Sergio Decherchi, Frank Xing, and Kenneth Kwok,
\newblock ``Senticnet 7: a commonsense-based neurosymbolic ai framework for
  explainable sentiment analysis,''
\newblock {\em Proceedings of LREC 2022}, 2022.

\bibitem{StanceSemEval2016}
Saif~M. Mohammad, Svetlana Kiritchenko, Parinaz Sobhani, Xiaodan Zhu, and Colin
  Cherry,
\newblock ``Semeval-2016 task 6: Detecting stance in tweets,''
\newblock in {\em Proceedings of the International Workshop on Semantic
  Evaluation}, San Diego, California, June 2016, SemEval '16.

\bibitem{liang2022jointcl}
Bin Liang, Qinlin Zhu, Xiang Li, Min Yang, Lin Gui, Yulan He, and Ruifeng Xu,
\newblock ``Jointcl: A joint contrastive learning framework for zero-shot
  stance detection,''
\newblock in {\em Proceedings of the 60th Annual Meeting of the Association for
  Computational Linguistics (Volume 1: Long Papers)}. Association for
  Computational Linguistics, 2022, vol.~1, pp. 81--91.

\bibitem{allaway2020zero}
Emily Allaway and Kathleen Mckeown,
\newblock ``Zero-shot stance detection: A dataset and model using generalized
  topic representations,''
\newblock in {\em Proceedings of the 2020 Conference on Empirical Methods in
  Natural Language Processing (EMNLP)}, 2020, pp. 8913--8931.

\bibitem{BERT}
Jacob Devlin, Ming{-}Wei Chang, Kenton Lee, and Kristina Toutanova,
\newblock ``{BERT:} pre-training of deep bidirectional transformers for
  language understanding,''
\newblock in {\em {NAACL-HLT}}. 2019, pp. 4171--4186, Association for
  Computational Linguistics.

\bibitem{liu2021enhancing}
Rui Liu, Zheng Lin, Yutong Tan, and Weiping Wang,
\newblock ``Enhancing zero-shot and few-shot stance detection with commonsense
  knowledge graph,''
\newblock in {\em Findings of the Association for Computational Linguistics:
  ACL-IJCNLP 2021}, 2021, pp. 3152--3157.

\bibitem{huang2023knowledge}
Hu~Huang, Bowen Zhang, Yangyang Li, Baoquan Zhang, Yuxi Sun, Chuyao Luo, and
  Cheng Peng,
\newblock ``Knowledge-enhanced prompt-tuning for stance detection,''
\newblock {\em ACM Transactions on Asian and Low-Resource Language Information
  Processing}, 2023.

\bibitem{LiangF00DHX21}
Bin Liang, Yonghao Fu, Lin Gui, Min Yang, Jiachen Du, Yulan He, and Ruifeng Xu,
\newblock ``Target-adaptive graph for cross-target stance detection,''
\newblock in {\em {WWW} '21: The Web Conference 2021, Virtual Event /
  Ljubljana, Slovenia, April 19-23}, 2021, pp. 3453--3464.

\bibitem{li2023tts}
Yingjie Li, Chenye Zhao, and Cornelia Caragea,
\newblock ``Tts: A target-based teacher-student framework for zero-shot stance
  detection,''
\newblock in {\em Proceedings of the ACM Web Conference 2023}, 2023, pp.
  1500--1509.

\end{thebibliography}

\end{document}